\documentclass[runningheads]{llncs}
\usepackage{graphicx}
\usepackage[table]{xcolor}
\usepackage{tikz}
\usepackage{comment}
\usepackage{amsmath,amssymb} 
\usepackage{color}

\usepackage[accsupp]{axessibility}  

\usepackage{epsfig}

\usepackage{url}            
\usepackage{booktabs}       
\usepackage{amsfonts}       
\usepackage{nicefrac}       
\usepackage{microtype}      
\usepackage{subfiles}
\usepackage{cite}
\usepackage{bm}
\usepackage{flexisym}
\usepackage{float}

\usepackage{pifont}
\newcommand{\cmark}{\text{\ding{51}}}
\newcommand{\xmark}{\text{\ding{55}}}

\usepackage{array}
\usepackage{multirow}
\newcolumntype{S}{>{\centering\arraybackslash}m{0.2cm}}
\newcolumntype{M}{>{\raggedright\arraybackslash}m{1.1cm}}
\newcolumntype{D}{>{\raggedright\arraybackslash}m{1.6cm}}
\newcolumntype{L}{>{\raggedright\arraybackslash}m{2.6cm}}
\newcolumntype{H}{>{\raggedright\arraybackslash}m{4.3cm}}
\definecolor{mygray}{gray}{.95}
\definecolor{mylightergray}{gray}{.99}
\definecolor{mygreen}{RGB}{10, 179, 33}
\usepackage{caption} 
\makeatletter
\newcommand{\thickhline}{%
    \noalign {\ifnum 0=`}\fi \hrule height 1pt
    \futurelet \reserved@a \@xhline
}
\newcolumntype{"}{@{\vrule width 1pt}}
\makeatother

\usepackage[ruled, lined, linesnumbered, commentsnumbered, longend]{algorithm2e}
\SetAlFnt{\footnotesize}
\SetAlCapFnt{\footnotesize}
\SetAlCapNameFnt{\footnotesize}
\SetKwInput{KwParam}{Parameter}
\SetKwInOut{KwIn}{Input}
\SetKwInOut{KwOut}{Output}

\SetCommentSty{mycommfont}
\SetAlgoCaptionSeparator{}

\newfloat{figtab}{htb}{fgtb}
\makeatletter
  \newcommand\figcaption{\def\@captype{figure}\caption}
  \newcommand\tabcaption{\def\@captype{table}\caption}
\makeatother

\usepackage[pagebackref,breaklinks,colorlinks]{hyperref}

\usepackage[capitalize]{cleveref}
\crefname{section}{Sec.}{Secs.}
\Crefname{section}{Section}{Sections}
\Crefname{table}{Table}{Tables}
\crefname{table}{Tab.}{Tabs.}

\usepackage{xspace}

\makeatletter
\DeclareRobustCommand\onedot{\futurelet\@let@token\@onedot}
\def\@onedot{\ifx\@let@token.\else.\null\fi\xspace}

\def\eg{\emph{e.g}\onedot} 
\def\ie{\emph{i.e}\onedot}

\def\etal{\emph{et al}\onedot}
\makeatother

\begin{document}
\pagestyle{headings}
\mainmatter
\def\ECCVSubNumber{4551}  

\title{VirtualPose: Learning Generalizable 3D Human Pose Models from Virtual Data} 

\titlerunning{VirtualPose}
%
\author{Jiajun~Su\inst{1, 2} \and
Chunyu~Wang\inst{5} \thanks{Corresponding author.} \and
Xiaoxuan~Ma\inst{2, 3} \and
Wenjun~Zeng\inst{6} \and
Yizhou~Wang\inst{2, 3, 4}}
\authorrunning{Su et al.}
%
\institute{Center for Data Science, Peking University \and 
Center on Frontiers of Computing Studies,
Peking University \and
Dept. of Computer Science, Peking University \and 
Inst. for Artificial Intelligence,
Peking University \and
Microsoft Research Asia \and
EIT Institute for Advanced Study
 \\
{\tt\small 
\{sujiajun, maxiaoxuan, yizhou.wang\}@pku.edu.cn, chnuwa@microsoft.com, zengw2011@hotmail.com}
}
\maketitle

\begin{abstract}
While monocular 3D pose estimation seems to have achieved very accurate results on the public datasets, their generalization ability is largely overlooked. In this work, we perform a systematic evaluation of the existing methods and find that they get notably larger errors when tested on different cameras, human poses and appearance. To address the problem, we introduce \emph{VirtualPose}, a two-stage learning framework to exploit the hidden ``free lunch'' specific to this task, \ie generating infinite number of poses and cameras for training models at no cost. To that end, the first stage transforms images to abstract geometry representations (AGR), and then the second maps them to 3D poses. It addresses the generalization issue from two aspects: (1) the first stage can be trained on diverse 2D datasets to reduce the risk of over-fitting to limited appearance; (2) the second stage can be trained on diverse AGR synthesized from a large number of virtual cameras and poses. It outperforms the SOTA methods without using any paired images and 3D poses from the benchmarks, which paves the way for practical applications. Code is available at \href{https://github.com/wkom/VirtualPose}{https://github.com/wkom/VirtualPose}.

\end{abstract}

\section{Introduction}
\label{sec:introduction}

\begin{figure}[ht]
	\centering
	\includegraphics[width=4.6in]{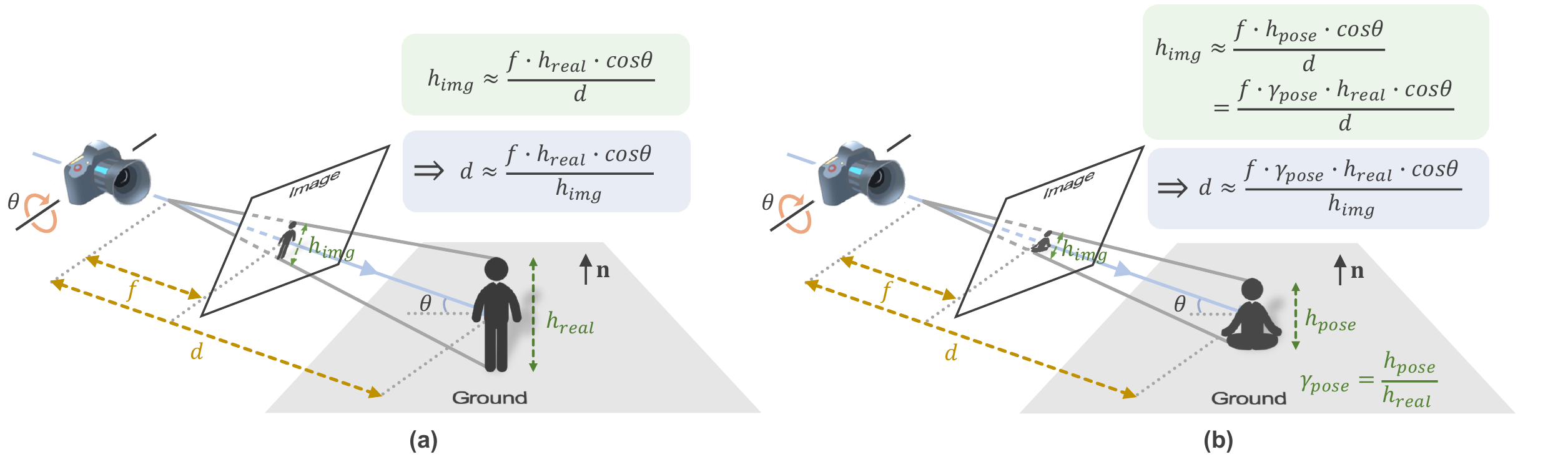}
	\caption{Projection geometry of a pinhole camera model. (a) shows the case where a person is in a standing pose. (b) shows the case with a different pose.}
	\label{fig:pinhole}
\end{figure}

Monocular 3D pose estimation has attracted much attention since it can benefit many applications. Most works \cite{ci2019optimizing,ma2021context,ci2020locally,wang2014robust,wang2018robust,martinez2017simple,moreno20173d,pavlakos2017coarse,sun2018integral} focus on a simpler sub-task of \emph{relative} 3D pose estimation where only relative joint positions are estimated. Absolute 3D pose estimation needs to estimate the depth of a person's \emph{root} joint in the camera coordinate system. This is more challenging because it is ill-posed and multiple entangled latent factors jointly determine the depth. As shown in Fig. \ref{fig:pinhole}, the relevant factors include at least the height of the person in neutral standing pose, its relative posture, its projection size in 2D, camera focal length, and camera view point. Some factors such as focal length may be assumed known in some cases but most others need to be implicitly estimated from images along with depth.

Many works \cite{zanfir2018monocular, zanfir2018deep,moon2019camera, lin2020hdnet, zhen2020smap, wang2020hmor, sarandi2020metrabs, cheng2022dual} propose to brute-forcely learn the mapping from images to depth.
Although they have got good results on the public datasets, they have poor generalization ability. We are aware that other tasks also face the issue but the situation is very different for pose estimation. First, the pose datasets \cite{sigal2010humaneva, ionescu2013human3, joo2015panoptic, mehta2017monocular, mehta2018single} have extremely limited variations in terms of cameras, human poses and appearance.  Second, many data augmentation techniques are not applicable, \eg, we can not change the view point neither the human poses in an image. So, addressing the generalization issue is non-trivial compared to other tasks. This was not identified as a serious issue previously because the current training and testing data \cite{ionescu2013human3, joo2015panoptic} are similar. In fact, even for the cross-dataset experiment (train on MuCo-3DHP \cite{mehta2018single} and test on MuPoTS-3D \cite{mehta2018single}), the camera views are also similar.

\begin{figure*}[t]
	\centering
	\includegraphics[width=4.8in]{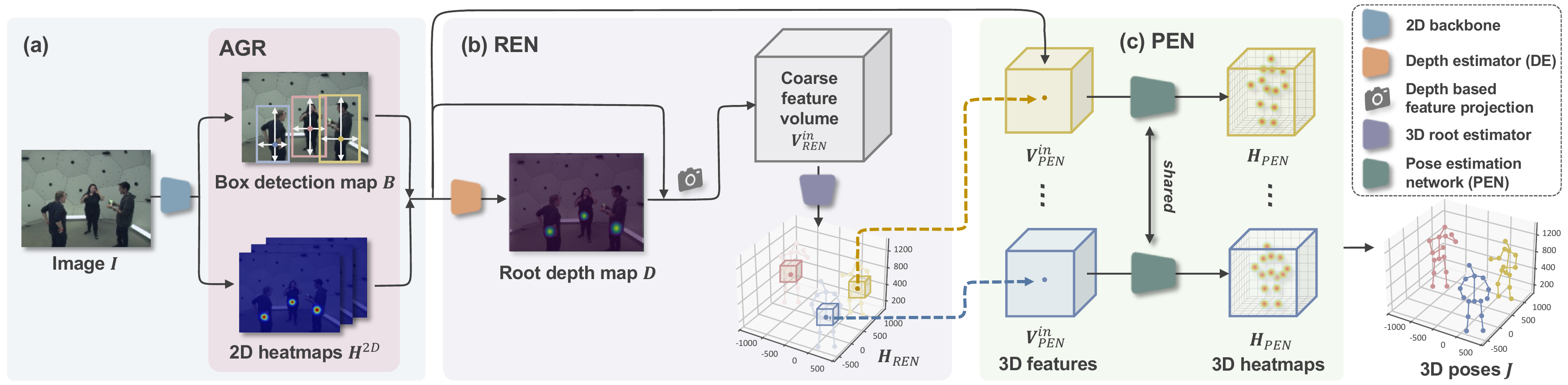}
	\caption{Our 3D human pose estimation pipeline. (a) Given an image as input, it first estimates the AGR (\ie bounding boxes and heatmaps) of all people. (b) Then, in Root Estimation Network (REN), it estimates the depth map of root joints from the AGR. The AGR and the depth map are integrated via depth based feature projection to estimate 3D root positions. (c) Finally, in Pose Estimation Network (PEN) for each root joint, we construct a feature volume around it to estimate a fine-grained 3D pose.}
	\label{fig:pipeline}
\end{figure*}

In this paper, we address the challenges by introducing an intermediate representation, termed as Abstract Geometry Representation (AGR), into the 3D pose estimation network. It is a bundle of multiple geometry representations that satisfy three requirements: (1) they are helpful for recovering absolute 3D poses, (2) can be synthesized from 3D poses, and (3) can be robustly estimated from images even when the model is trained on wild images rather than mocap datasets. As shown in Fig. \ref{fig:pipeline}, AGR splits a 3D pose estimator into two successive modules. The first module maps a raw image to AGR which, in current implementation, consists of human detection and joint localization results. Since it only handles 2D tasks which are barely affected by 3D projection geometry, we can train the model on the diverse 2D datasets such as COCO \cite{lin2014microsoft}, which covers a large number of camera views, human poses and backgrounds, and apply extensive data augmentation. As a result, the module is robust to different factors and achieves desirable results on wild data.

The second module learns to regress 3D pose from AGR. Different from the previous works, we propose a novel training strategy which synthesizes a large number of paired $<$AGR, Pose$>$ data from diverse camera views, poses and person positions to learn a generalizable model. It is worth noting that AGR suffers less from domain gaps than raw images. In some cases when we want to deploy a model for a fixed environment, \eg installing a camera at home for elderly care, we can even generate virtual training data specifically for the environment, which as shown in our experiment, gets more accurate results. Combining the two modules, we get an accurate yet generalizable 3D pose estimator. It not only outperforms the state-of-the-art methods on the benchmark datasets but also achieves good results on our own videos collected in retail stores with cluttered background and severe occlusion.

We implement the above idea following the architecture of VoxelPose \cite{tu2020voxelpose}, as shown in Fig. \ref{fig:pipeline}. Given an image as input, we first estimate bounding boxes and 2D pose heatmaps as AGR using a CNN network. Then they are integrated into the 3D space to estimate the 3D positions of all persons in the image.  Finally, for each person, we construct a feature volume around its position to estimate a fine-grained 3D pose. 

In summary, our contributions are three-fold. First, we present the first systematic study on models' generalization ability  which is largely overlooked in the previous works. We argue that it is important to purposely prevent models from over-fitting when designing and evaluating new models. Second, we present one possible way for learning generalizable models purely from virtual 3D poses and cameras which is made possible by AGR. Note that the decoupling strategy has been used in previous works \cite{chang2019absposelifter, tu2020voxelpose,wu20183d,von2018recovering, li2020cascaded, pavllo20193d} for different motivations. Our contribution lies in leveraging it in our training strategy for absolute 3D pose estimation with concrete designs. Finally, the method outperforms the existing methods without using paired images and 3D poses for training, which notably improves its applicability in practice.

\section{Related Work}
\label{sec:related work}

\paragraph{\bf Datasets and Generalization} Collecting motion capture datasets is cumbersome because it either uses multiview systems \cite{sigal2010humaneva, ionescu2013human3, joo2015panoptic, mehta2017monocular, mehta2018single} or wearable IMUs \cite{von2018recovering,zhang2020fusing}. As a result, they usually have limited camera views, poses and actors. Some works propose to generate more data by changing the image backgrounds \cite{mehta2018single, moon2019camera} of the existing datasets, or using game engines \cite{zhu2020reconstructing, cao2020long,zhang2021adafuse}. However, the domain gap between synthetic and real images poses challenges for neural networks. We find that the models trained on these datasets lack generalization capability. Some works \cite{zhou2017towards, yang20183d, wandt2019repnet} address the generalization problem in relative pose estimation but they cannot be extended to the absolute task. There are some works in other areas that also decouple a network into different stages \cite{wu20183d,tu2020voxelpose,von2018recovering}. While we also follow this general idea, we have specific designs to apply it to the 3D absolute pose estimation task.

\paragraph{\bf Minimizing Projection Error} The methods of this class \cite{rogez2017lcr, rogez2019lcr,dabral2019multi, fabbri2020compressed} first estimate relative 3D human poses with sizes from an image in an end-to-end manner. Then based on the estimated 2D projection size and the focal length, they can coarsely compute the depth of each pose by minimizing an error between the 2D pose and the projection of the estimated 3D pose. The results of these methods heavily rely on the accuracy of relative 3D pose estimation. Some later works \cite{zhou2017towards,wandt2019repnet} show that they have poor generalization results because they need to be trained on the (small) motion capture datasets \cite{ionescu2013human3, mehta2017monocular}.

\paragraph{\bf Ground Plane Geometry} Some works such as \cite{mehta2019xnect} propose to analytically compute the real 3D size and depth from ground plane geometry. They assume that the camera is calibrated, people are standing on the ground and their feet are visible in the image. Then the absolute human height and depth can be calculated by projection geometry. However, these assumptions are not always true in practice which limits their applicability.

\paragraph{\bf End-to-end Learning} Some works \cite{moon2019camera, lin2020hdnet, zhen2020smap, wang2020hmor} directly estimate depth from images. We first explain why this is possible. Consider a simple case where a person with height $h_{real}$ stands on the ground plane and a camera with focal length $f$ is placed with pitch angle $\theta$, as shown in Fig. \ref{fig:pinhole}(a). Then the height of the person in the image $h_{img}$ is approximately $h_{img} \approx \frac{f \cdot h_{real} \cdot cos \theta}{d}$. In general, people may be in various poses, \eg sitting on the ground as shown in Fig. \ref{fig:pinhole}(b). We define the height of the person perpendicular to the ground as $h_{pose}$, and a pose-dependent correction factor as $\gamma_{pose} = \frac{h_{pose}}{h_{real}}$, then we have $h_{img} \approx \frac{f \cdot \gamma_{pose} \cdot h_{real} \cdot cos \theta}{d}$, from which the depth $d$ of the person can be computed. 

These methods usually assume $f$ is known and estimate normalized depth $d^{norm} = \frac{d}{f}$. It is challenging to estimate $h_{real}$ from a single image. But since there are only few actors in the benchmark datasets, they are implicitly learned by the network. The rest factors of $\gamma_{pose}$, $\theta$ and $h_{img}$ can be estimated from images when the training data is large and diverse.
However, since most benchmark datasets are small, \eg having limited camera poses, human poses and backgrounds, the depth estimation models may easily over-fit to these datasets but can not generalize well on unseen images.

\section{Generalization Study}
\label{sec:factor}

In this section, we systematically evaluate the robustness of the existing methods to the variations of the key factors. In Section \ref{subsec:baseline}, we first introduce three representative baselines for estimating depth which are most frequently used by the existing 3D pose estimators \cite{moon2019camera, guo2021monocular,lin2020hdnet,veges2019absolute, zhen2020smap,wang2020hmor,zhou2019objects}. Then we evaluate their robustness to the three factors including camera pose, image background and human pose.

\subsection{Baselines and Datasets}
\label{subsec:baseline}

\paragraph{\bf Top-down Box Size based method (TBS)} Given an image as input, it first localizes all people by bounding boxes and then estimates depth for each people. They assume that the 3D sizes of all people are constant and known, and geometrically compute the depth based on the 2D size of the bounding box. However, as discussed in Fig. \ref{fig:pinhole}(b), the 2D size is also dependent on the human pose. So they learn a correction factor from images. The factor is used to refine the estimated depth.
Multiple methods \cite{moon2019camera, guo2021monocular} follow this strategy. We choose RootNet \cite{moon2019camera} in our experiments.

\paragraph{\bf Top-down Image Feature based method (TIF)} It first detects all people in the input image. But instead of using the size of the detected bounding box to analytically compute depth, they use the estimated 2D heatmaps as attention masks to pool features from the image and directly predict the human depth. We choose HDNet \cite{lin2020hdnet} in our experiments.

\paragraph{\bf Bottom-up Depth Regression based method (BDR)} This method directly estimates a depth map from each image using a deep network. The depth of each joint can then be obtained from the depth map directly. Many works \cite{veges2019absolute, zhen2020smap, wang2020hmor} follow this pipeline. We present a simple 3D pose estimator which has two branches for estimating 2D pose heatmaps and depth maps, respectively. The 3D pose can then be analytically computed from them.

\paragraph{\bf Implementation Details} 
In order to mitigate the influence of detection results, we use the GT human bounding box in TBS and TIF methods. The backbone in the BDR method is ResNet-50 \cite{he2016deep}. In order to mitigate the influence of 2D pose estimation, we use the GT 2D root position to obtain root depth from the depth maps in BDR method.  We conduct experiments on the CMU Panoptic \cite{joo2015panoptic}, MuCo-3DHP and MuPoTS-3D \cite{mehta2018single} datasets.

\paragraph{\bf CMU Panoptic}\cite{joo2015panoptic} is a large-scale multi-person pose dataset captured in a studio with multiple cameras. We select data of four activities (Haggling, Mafia, Ultimatum, Pizza) as our training and test set. The training and testing splits are different in different ablative experiments and will be made clear as needed.

\begin{figure}[t]
	\centering
	\includegraphics[width=4.8in]{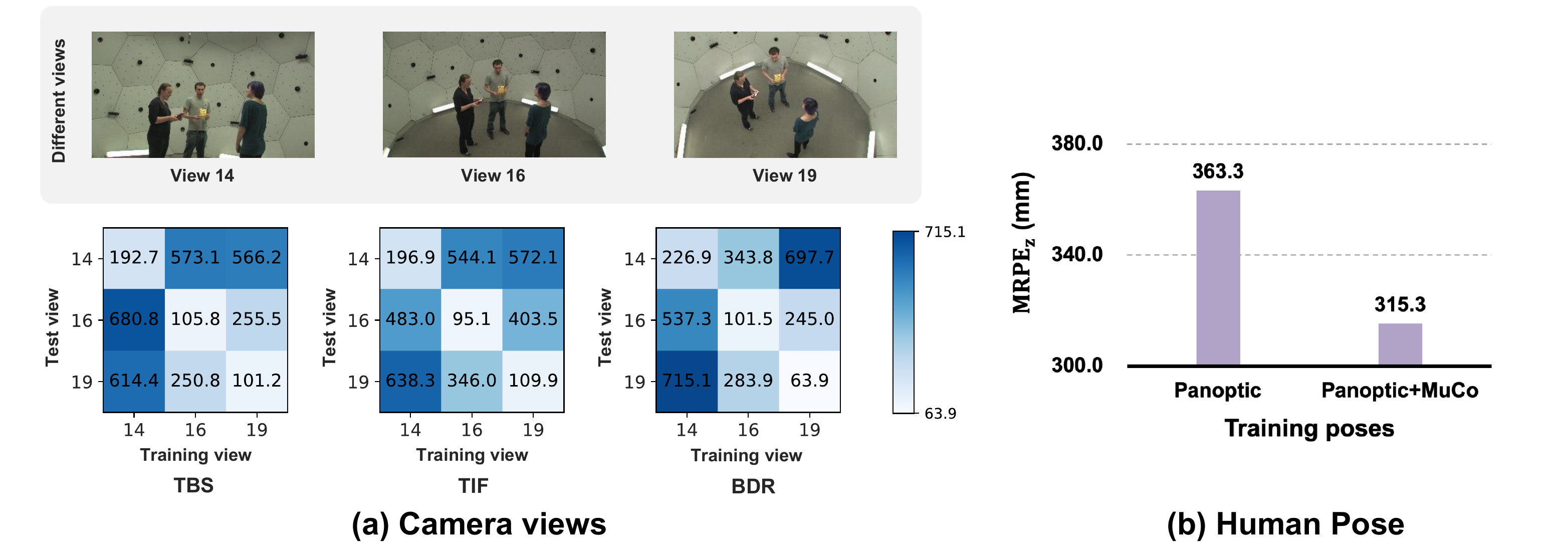}
    \caption{(a) Cross-view results ($\rm MRPE_z$) of the three models on the CMU Panoptic dataset. The x- and y-axes represent the camera index in training and test, respectively. (b) Depth estimation results ($\rm MRPE_z$) when the training poses are from the CMU Panoptic dataset, or from the mixture of ``Panoptic+MuCo'' datasets. The MuPoTS-3D dataset is the test dataset.}
	\label{fig:comparison_camera_pose}
\end{figure}

\paragraph{\bf MuCo-3DHP}\cite{mehta2018single} is created by compositing randomly sampled 3D poses from single-person 3D human pose dataset MPI-INF-3DHP \cite{mehta2017monocular} to synthesize multi-person scenes. The images are synthesized by resizing and compositing the corresponding segmented images of each person according to their 3D positions. We use the data provided by \cite{moon2019camera} which contains images with green screen background and images with augmented background using COCO \cite{lin2014microsoft} dataset.

\paragraph{\bf MuPoTS-3D}\cite{mehta2018single} is a multi-person test set comprising 20 general real world scenes with GT 3D pose obtained with a multi-view marker-less motion capture system. It contains 5 indoor and 15 outdoor settings, varying human poses and camera poses, which is a commonly used test set to validate the models' generalization ability.

\begin{table}
    \centering
    \caption{Depth estimation results ($\rm MRPE_z$) when we train the models on images with green-screen background (green) or with mixture of green-screen and augmented background (augmented), respectively. }
    \setlength{\tabcolsep}{3.5mm}{
    \begin{tabular}{l|l|cc} 
    \thickhline 
    \multirow{2}{*}{Method} & \multirow{2}{*}{Training data} & \multicolumn{2}{c}{Test data} \\
    \cline{3-4}
    & & MuCo (green) & MuCo (augmented) \\
    \hline
    \rowcolor{mygray}
                             & green & 76.0 & 281.6 \\
    \rowcolor{mygray}
    \multirow{-2}{*}{TBS} & augmented & 73.3 & 82.5 \\
    \multirow{2}{*}{TIF} & green  & 57.3 & 168.8 \\
                           & augmented & 58.7 & 65.5\\
    \rowcolor{mygray}
                               & green & 126.3 & 267.8 \\
    \rowcolor{mygray}
    \multirow{-2}{*}{BDR} & augmented & 141.7 & 148.9 \\
    \thickhline
    \end{tabular}}
    \label{tab:comparison_background}
\end{table}

\paragraph{\bf Metrics} We focus on evaluating the quality of the estimated depth of root joint in this section.  We use the metric of $\rm MRPE_z$ (in mm) proposed in \cite{moon2019camera}, which is the mean of the root position errors of all people in the $z$ direction (\ie the depth direction in the camera coordinate system).

\subsection{Experimental Results}
\label{subsec:factor_experiment}
\paragraph{\bf Camera Views} We conduct the cross-view experiment on the CMU Panoptic \cite{joo2015panoptic} dataset using cameras of $14$, $16$ and $19$ as shown in Fig. \ref{fig:comparison_camera_pose}(a). We train the three baseline models on one of the cameras and test them on the rest. The results are shown in Fig. \ref{fig:comparison_camera_pose}(a). All methods get several times larger depth estimation errors when they are trained and tested on different cameras. The results are reasonable because (1) camera angle affects image appearance and image features; (2) camera angle also affects the 2D size of the people in the image. The results suggest that the models learned on few camera views are barely generalizable. It is helpful to train the models on as many views as possible.

\paragraph{\bf Image Background} For each method, we train two models on the MuCo-3DHP \cite{mehta2018single} dataset, with the first on images with green screen background and the second with both green screen and augmented background. We select $70\%$ of the dataset for training, and the rest for testing. Tab. \ref{tab:comparison_background} shows the results. When tested on the green screen subset of MuCo-3DHP, BDR (green) achieves a smaller error than BDR (augmented). It means that BDR probably learns some scene priors to reduce the ambiguity in estimation but it may also risk over-fitting. For example, BDR (green) gets a significantly larger error on the augmented subset of MuCo-3DHP. The results not only validate the limitations of those methods on small training datasets, but also suggest that we actually need a better evaluation protocol, \eg training and testing on datasets with large differences, to convincingly evaluate the future works.

\paragraph{\bf Human Pose} We train two models either using poses only from the CMU Panoptic dataset, or from the combined CMU Panoptic and MuCo-3DHP datasets. We test the models on the MuPoTS-3D dataset. To keep other factors such as human appearance the same, we modify BDR to take pose heatmaps as input and output the root depth map $\bm{D}$ (this is actually part of our method as will be described). The results are shown in Fig. \ref{fig:comparison_camera_pose}(b). The model trained on CMU Panoptic poses achieves a larger error meaning that the model trained on limited poses may have poor generalization results. While the conclusion is not surprising, it actually points out an overlooked problem that relying on limited mocap datasets 
to train 3D pose estimators will probably fail in real-world applications.

\section{VirtualPose}
\label{sec:method}
In this section, we present our \emph{VirtualPose}. To address the challenges, we introduce an intermediate representation, termed as Abstract Geometry Representation (AGR, Section \ref{subsec:intermediate_representation}), into the 3D pose estimation network so that we can simultaneously leverage the diverse 2D datasets and large amounts of synthetic data. As shown in Fig. \ref{fig:pipeline}, AGR splits a 3D pose estimator into two successive modules. The first module maps a raw image to AGR which consists of human detection and joint localization results.
Then the second module maps AGR to the corresponding 3D pose which is trained on synthesized $<$AGR, Pose$>$ data. In particular, we present Root Estimation Network (REN, Section \ref{subsec:root_estimation_network}) 
to estimate 3D positions of the persons. Then for each person, Pose Estimation Network (PEN, Section \ref{subsec:pose_estimation_network}) is proposed to estimate the 3D pose.

\subsection{Abstract Geometry Representation}
\label{subsec:intermediate_representation}
AGR is a general concept representing a bundle of geometry representations that satisfy three requirements: (1) they are helpful for recovering absolute 3D poses, and (2) can be robustly estimated from images even when the model is only trained on the wild 2D datasets rather than the 3D motion capture datasets, and (3) can be synthesized or rendered from 3D poses. In current implementation, we only use 2D pose heatmaps and human detection bounding boxes in AGR for simplicity. But more cues such as ordinal depth, occlusion relationship or segmentation mask can also be leveraged. 

We adopt a simple architecture for estimating AGR from images following CenterNet \cite{zhou2019objects}.
As shown in Fig. \ref{fig:pipeline}(a), given an image $\bm{I}$ as input, the 2D backbone outputs the corresponding 2D pose heatmaps $\bm{H}^{2D} \in [0, 1]^{W \times H \times N}$ and box detection map $\bm{B} \in \mathcal{R}^{W \times H \times 4}$ where the four channels encode the distances from the human root joint to the four edges of the human bounding box, respectively. Here, $N$ indicates the number of human joints, and $W, H$ are the width and height of the output maps. The 2D human pose estimation and detection network is trained by:
\begin{equation}
\label{eq:2D_loss}
\begin{aligned}
    \mathcal{L}_{2D} = \mathcal{L}^{2D}_{heat} + \lambda_{bbox} \mathcal{L}_{bbox}
\end{aligned}
\end{equation}
\begin{equation}
\label{eq:2Dheat_loss}
\begin{aligned}
    \mathcal{L}^{2D}_{heat} = ||\bm{H}^{2D} - \widetilde{\bm{H}}^{2D}||_2^2
\end{aligned}
\end{equation}
\begin{equation}
\label{eq:bbox_loss}
\begin{aligned}
    \mathcal{L}_{bbox} = \sum_{p \in \mathcal{P}}||\bm{B}_{p} - \widetilde{\bm{B}}_{p}||_1,
\end{aligned}
\end{equation}
where $\widetilde{\bm{H}}^{2D}$ is the GT 2D pose heatmaps and $\lambda_{bbox}$ is a hyper-parameter. The box supervision is only enforced at GT root joint locations $\mathcal{P}$ where $\bm{B}_{p}$ is the estimated box embedding for the $p_{th}$ person and $\widetilde{\bm{B}}_{p} \in \mathcal{R}^4$ is the GT box embedding.



\subsection{Root Estimation Network}
\label{subsec:root_estimation_network}
We first estimate a coarse depth map for the root joint as shown in Fig. \ref{fig:pipeline}, which will be used to reduce the ambiguity when projecting 2D heatmaps to the 3D space. To that end, the previously estimated 2D pose heatmaps $\bm{H}^{2D}$ are fed to the 2D CNN-based Depth Estimator (DE) to estimate a root depth map $\bm{D}$. 
It is trained by minimizing:
\begin{equation}
\label{eq:depth_loss}
 \small
\begin{aligned}
    \mathcal{L}_{depth} = \sum_{p \in \mathcal{P}}|\bm{D}_{p} - \widetilde{\bm{D}}_{p}|,
\end{aligned}
\end{equation}
where $\bm{D}_{p}$ and $\widetilde{\bm{D}}_{p}$ are the estimated and GT depth values of the root joint for the $p_{th}$ person.

Then we estimate the 3D positions of the root joints by constructing a 3D representation. We discretize the 3D space in which people can freely move into $X \times Y \times Z$ voxels $\{\bm{G}_{x, y, z}\}$, and construct a 3D feature volume $\bm{V}_{REN}^{in} \in [0,1]^{X \times Y \times Z \times N}$ by inversely projecting 2D pose heatmaps of $N$ body joints to the 3D space using camera parameters. Different from the previous works \cite{ tu2020voxelpose,zhang2022voxeltrack}, for a heatmap vector at a pixel location within a bounding box, we only project it to a few voxels whose depths are similar to the estimated depth $\bm{D}_{p}$ of the root joint in the bounding box. For example, if a voxel $\bm{G}_{x, y, z}$ whose projected 2D pixel location $(u, v)$ is within the $p_{th}$ human bounding box, then the feature of $\bm{G}_{x, y, z}$ can be calculated as
\begin{equation}
\label{eq:project}
\small
\begin{aligned}
       \bm{V}_{REN}^{in}(x, y, z) = \bm{H}^{2D}(u, v)  \exp(-\frac{(z -  \bm{D}_{p})^2}{2\sigma^2}),
\end{aligned}
\end{equation}
where $\sigma$ is empirically set to be $200$. This is particularly important for monocular 3D pose estimation which helps reduce the ambiguity. The feature volume $\bm{V}_{REN}^{in}$ coarsely encodes the likelihood of each person's position in each voxel. We feed $\bm{V}_{REN}^{in}$ to a 3D network to estimate the corresponding 3D heatmaps $\bm{H}_{REN} \in [0,1]^{X \times Y \times Z}$ indicating the confidence of each voxel containing a root joint as shown in Fig. \ref{fig:pipeline}(b). We select the voxels with large confidence values in the estimated $\bm{H}_{REN}$ with Non-Maximum Suppression (NMS). We train the 3D network by minimizing:
\begin{equation}
\label{eq:position_loss}
\begin{aligned}
    \mathcal{L}_{REN} = ||\bm{H}_{REN} - \widetilde{\bm{H}}_{REN}||_2^2,
\end{aligned}
\end{equation}
where $\widetilde{\bm{H}}_{REN} \in [0, 1]^{X \times Y \times Z}$ is the GT 3D heatmap for root joint. We set $X$, $Y$ and $Z$ to be $80$, $80$ and $24$, respectively.

\subsection{Pose Estimation Network}
\label{subsec:pose_estimation_network}

After estimating the root joint position for each person, we will estimate a complete 3D pose for it. As shown in Fig. \ref{fig:pipeline}(c), we first construct a finer-grained feature volume centered at the estimated root joint position by inversely projecting the estimated 2D pose heatmaps $\bm{H}^{2D}$. The spatial size of the feature volume $\bm{V}_{PEN}^{in} \in [0, 1]^{X' \times Y' \times Z' \times N}$ is set to be $2m \times 2m \times 2m$ which is sufficiently large to cover people in arbitrary poses. We set $X' = Y' = Z' = 64$, so approximately each voxel has a size of $30$mm. 

We feed $\bm{V}_{PEN}^{in}$ to a 3D Pose Estimation Network (PEN) to estimate 3D heatmaps $\bm{H}_{PEN} \in [0, 1]^{X' \times Y' \times Z' \times N}$ for $N$ joints, including the root joint. For each joint $k$, the 3D location $\bm{J}_k$ can be obtained using the integration trick proposed in \cite{sun2018integral} to reduce quantization error:
\begin{equation}
\label{eq:integral}
\begin{aligned}
       \bm{J}_k = \sum_{x=1}^{X'}\sum_{y=1}^{Y'}\sum_{z=1}^{Z'}(x, y, z) \cdot \bm{H}_{PEN, k}(x, y, z).
\end{aligned}
\end{equation}

We train PEN using the $L_1$ loss as in \cite{tu2020voxelpose}:

\begin{equation}
\label{eq:pose_loss}
\begin{aligned}
       \mathcal{L}_{PEN} = \frac{1}{N} \sum^{N}_{k=1} ||\bm{J}_k - \widetilde{\bm{J}}_k||_1.
\end{aligned}
\end{equation}
PEN has the same network structure as the 3D network in REN, and the weights are shared for different people. We find that by estimating the absolute 3D locations of all joints, it further improves the absolute position of the root joint compared to the REN output.










\label{subsec:training}

\section{Experiments}
\label{sec:experiments}

\subsection{Implementation Details}
\label{subsec:implementation}

\paragraph{\bf General Details}
The 2D backbone for estimating boxes and 2D pose heatmaps is ResNet-152 \cite{he2016deep}. The backbone for estimating depth map from heatmaps is ResNet-18 \cite{he2016deep}. The input image is resized to $512 \times 960$ and the size of the resulting heatmap is $128 \times 240$. We adopt Adam \cite{kingma2015adam} as optimizer, the learning rate is $1 \times 10^{-4}$ and the batch size is $24$. The 2D backbone network is trained for $40$ epochs. The depth estimator, REN and PEN are trained end-to-end with synthetic heatmaps and their GT 3D poses for $20$ epochs.

\paragraph{\bf Training Data}
For the experiments on CMU Panoptic\cite{joo2015panoptic}, the backbone is trained on the combined COCO\cite{lin2014microsoft} and CMU Panoptic datasets. When synthesizing AGR training data, the absolute 3D poses and camera views are the same as the original training data in Panoptic dataset. For the MuPoTS-3D\cite{mehta2018single} experiment, the backbone is trained on the combined COCO and MuCo-3DHP\cite{mehta2018single} datasets. When synthesizing AGR training data, the relative 3D poses and camera views are from MuCo-3DHP, and we randomly place the poses in 3D space to enhance the diversity of the training data.

\begin{table}
    \centering
    \caption{Comparison to the state-of-the-art methods on the CMU Panoptic dataset. The metric is MPJPE (mm). \cite{wang2020hmor}$^{\dagger}$ uses the GT depth when estimating absolute 3D poses so it is not fairly comparable to other methods. Our method does not use paired images and poses data for training but it still achieves smaller errors than other methods.}
    \setlength{\tabcolsep}{2mm}{
    \begin{tabular}{l|ccccc} 
    \thickhline 
    Method & Haggling &  Mafia & Ultimatum & Pizza & Mean $\downarrow$ \\
    \hline
    \rowcolor{mygray}
    Li \etal (\textbf{HMOR}) \cite{wang2020hmor}$^{\dagger}$ & 50.9 & 50.5 & 50.7 & 68.2 & 55.1 \\
    \hline
    PoPa \etal \cite{popa2017deep} & 217.9 & 187.3 & 193.6 & 221.3 & 203.4 \\
    \rowcolor{mygray}
    Zanfir \etal \cite{zanfir2018monocular} &  140.0 & 165.9 & 150.7 & 156.0 & 153.4 \\
    Moon \etal (\textbf{RootNet}) \cite{moon2019camera} &  89.6 & 91.3 & 79.6 & 90.1 & 87.6 \\
    \rowcolor{mygray}
    Zanfir \etal \cite{zanfir2018deep} &  72.4 & 78.8 & 66.8 & 94.3 & 78.1 \\
    Zhen \etal (\textbf{SMAP}) \cite{zhen2020smap} & \underline{63.1} & \textbf{60.3} & \underline{56.6} & \underline{67.1} & \underline{61.8} \\
    \hline
    \rowcolor{mygray}
    Ours & \textbf{54.1} & \underline{61.6} & \textbf{54.6} & \textbf{65.4} & \textbf{58.9} \\
    \thickhline
    \end{tabular}}
    \label{tab:sota_panoptic}
\end{table}

\subsection{Comparison to the State-of-the-arts}
\label{subsec:comparision}

\paragraph{\bf CMU Panoptic}\cite{joo2015panoptic} Following \cite{zanfir2018monocular}, we use the video sequences of camera $16$ and $30$ as training and test data which consist of four activities, \ie Haggling, Mafia, Ultimatum, Pizza. We show the results in Table ~\ref{tab:sota_panoptic}. We achieve better results than previous methods except for HMOR \cite{wang2020hmor}  which uses the GT depth when estimating the absolute 3D pose. Besides, our method achieves the best performance in the Pizza sequence which is not included in the training data. This partially validates the generalization ability of our method.

\paragraph{\bf MuPoTS-3D}\cite{mehta2018single} \
Our method is trained on the synthetic data where the 3D poses are from the MuCo-3DHP dataset. The rest methods are trained end-to-end on paired images and 3D poses from the MuCo-3DHP dataset. Following the standard practice on this dataset, the metric of percentage of correct keypoints (3DPCK) is used to measure the estimation results.  The results are shown in Table \ref{tab:sota_mupots}. $\rm PCK_{abs}$ measures the accuracy of absolute pose and $\rm PCK_{root}$ measures the accuracy of the root joint. We can see that our method achieves significantly better depth and pose estimation results than the state-of-the-arts. It validates that our absolute 3D depth estimation method has strong generalization capability. We hope to emphasize the importance of the results because it means the method has the potential to be applied in the wild environments.

\begin{table}[t]
    \centering
    \caption{Comparison to the state-of-the-art methods on MuPoTS-3D. \emph{Matched people} only computes accuracy for GT poses which are matched to predictions and \emph{All people} computes accuracy for all GT poses in the dataset. The methods are not strictly comparable because they may have different backbones.}
    \setlength{\tabcolsep}{3.8mm}{
    \begin{tabular}{l|cc|c} 
    \thickhline 
    \multirow{2}{*}{Method} & \multicolumn{2}{c|}{Matched people} &  \multicolumn{1}{c}{All people} \\
    \cline{2-4}
    & $\rm PCK_{abs} \uparrow$ & $\rm PCK_{root} \uparrow$ & $\rm PCK_{abs} \uparrow$ \\
    \hline
    \rowcolor{mygray}
    Moon \etal (\textbf{RootNet}) \cite{moon2019camera} & 31.8 & 31.0 & 31.5 \\
    Lin \etal (\textbf{HDNet}) \cite{lin2020hdnet} & 35.2 & 39.4 & - \\
    \rowcolor{mygray}
    Zhen \etal (\textbf{SMAP}) \cite{zhen2020smap} & 38.7 & 45.5 &  35.4 \\
    Veges \etal \cite{veges2020multi} & 39.6 & - & 37.3 \\
    \rowcolor{mygray}
    Sarandi \etal \cite{sarandi2020metrabs} & 40.5 & - & 38.4 \\
    Li \etal (\textbf{HMOR}) \cite{wang2020hmor} & - & - & 43.8 \\
    \rowcolor{mygray}
    Guo \etal \cite{guo2021monocular} & 39.6 & - & 39.2 \\
    
    \hline
    
    Ours & \textbf{47.0} & \textbf{53.5} & \textbf{44.0} \\

    \thickhline
    \end{tabular}}
    \label{tab:sota_mupots}
\end{table}

\subsection{Ablation Study}
\label{subsec:ablation}

In this section, we will evaluate the impact of our proposed modules and the training strategies to the estimation results of the root joint.

\paragraph{\bf Root Estimation Network} We first introduce two baselines. Baseline (a) analytically computes the 3D root position based on the 2D position and the estimated depth map. The baseline (b) uses the 3D root estimator to estimate the 3D root position on top of (a). The results on the CMU Panoptic and MuPoTS-3D datasets are shown in Table \ref{tab:ablation_effect}. By comparing the results of (a) and (b), we can see that the root depth error has a significant reduction which validates the effectiveness of the 3D root estimator.

\paragraph{\bf Pose Estimation Network}
By comparing the results of (b) and (c) in Table \ref{tab:ablation_effect}, we can see that depth estimation can be notably improved by PEN, which leverages the rest of the body joints to refine the root joint. Another reason for the improvement is that the quantization error is reduced by computing continuous root locations via the integral trick.

\paragraph{\bf Depth Based Feature Projection} As stated in Section \ref{subsec:root_estimation_network}, we construct the 3D feature volume by projecting 2D pose heatmaps based on the estimated depths and bounding boxes. We compare it to a baseline which naively projects the heatmaps to all voxels as in \cite{iskakov2019learnable, tu2020voxelpose} and find that $\rm MRPE$ and $\rm MRPE_z$ of our method are significantly better than the baseline ($204.0$mm \emph{vs.} $253.2$mm, and $159.8$mm \emph{vs.} $210.5$mm) on the MuPoTS-3D dataset.

\begin{table}
    \centering
    \caption{Ablation study on Root Estimation Network (REN) and Pose Estimation Network (PEN) in our method. We report the $\rm MRPE$ and $\rm MRPE_z$ (mm) on the test set of CMU Panoptic and MuPoTS-3D dataset. }
    \setlength{\tabcolsep}{2mm}{
    \begin{tabular}{l|c|c|cc|cc} 
    \thickhline 
    \multirow{2}{*}{Method} & \multirow{2}{*}{REN} & \multirow{2}{*}{PEN} & \multicolumn{2}{c|}{CMU Panoptic} & \multicolumn{2}{c}{MuPoTS-3D} \\
    \cline{4-7}
    & & & $\rm MRPE \downarrow$ & $\rm MRPE_z \downarrow$ & $\rm MRPE \downarrow$ & $\rm MRPE_z \downarrow$  \\
    \hline
    \rowcolor{mygray}
    (a) DE & 2D & \textcolor{red}{\xmark} & 113.9  & 104.1 &  282.0 & 245.4  \\	
    (b) REN & 2D+3D & \textcolor{red}{\xmark} & 115.7 & 93.6  & 217.5 & 165.0   \\	\rowcolor{mygray}
    (c) Ours & 2D+3D & \textcolor{mygreen}{\cmark} &\textbf{97.0} & \textbf{86.0} &  \textbf{204.0} & \textbf{159.8}  \\
    \thickhline
    \end{tabular}}
    \label{tab:ablation_effect}
\end{table}

\paragraph{\bf Training Data Generation Strategies} 
We compare several training data generation strategies for different scenarios. First, when we have little knowledge about the testing camera view point, we can only uniformly sample camera views for the virtual camera to synthesize training data. The results are shown in Table \ref{tab:experiment_random_view}. We can see that by training a universal model, our method achieves smaller depth estimation errors than the model trained on a single camera.

Second, if we know the camera view point in the testing environment, we can generate training data specifically for it. If we train a model for camera 19, then the testing error on camera 19 will be decreased significantly to $62.3$mm. This is helpful when the camera is fixed, \eg deployed at home for child or elderly care. In this case, we can get very accurate estimation results. We have similar observations for the pose factor. In particular, if we have the pose information for the people who are going to appear in the camera, then we can generate AGR using those poses. In that case, the testing accuracy will also be significantly improved. For example, if we train the model using the poses from the MuPoTS-3D dataset, then the $\rm PCK_{abs}$ will be improved from $44.0\%$ to $50.4\%$.

\begin{table}
    \centering
    \caption{Impact of training data generation strategies. The metric is $\rm MRPE_{z}$ (mm). We can see that by training a universal model, our method achieves smaller depth estimation errors than the model trained on a single camera.}
    \setlength{\tabcolsep}{2mm}{
    \begin{tabular}{l|c|c|c} 
    \thickhline 
     \multirow{2}{*}{Training Camera View} & \multicolumn{3}{c}{Test Camera view} \\
    \cline{2-4}
    & 14 & 16 & 19 \\
    \hline
    \rowcolor{mygray}
    14 & - & 170.3 & 294.6 \\
    16 & 454.6 & - & 198.6 \\
    \rowcolor{mygray}
    19 & 621.3 & 385.1 & - \\
    \hline
    Random & \textbf{273.8} & \textbf{134.0} & \textbf{155.2} \\
    \thickhline
    \end{tabular}}
    \label{tab:experiment_random_view}
\end{table}

\subsection{Qualitative Results}
\label{subsec:qualitative}

\begin{figure}[t]
	\centering
	\includegraphics[width=5in]{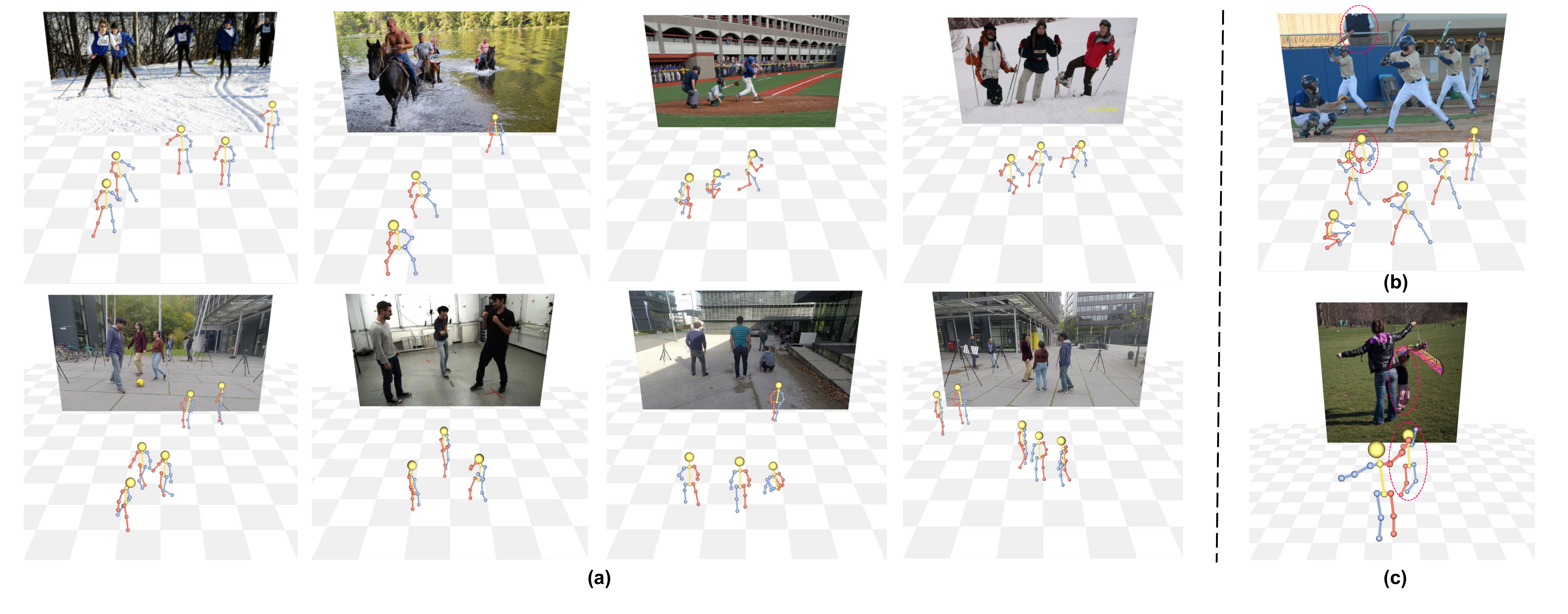}
    \caption{(a) Some pose estimation results on the COCO (top) and MuPoTS-3D (bottom) datasets. (b) Typical failure cases due to occlusion and truncation. (c) Typical failure cases due to the person is shorter than those in the training dataset. As a result, the estimated depth is larger than GT.}
	\label{fig:experiment_quality}
\end{figure}

Fig. \ref{fig:experiment_quality}(a) shows some estimation results for images from the COCO dataset and MuPoTS-3D dataset. Since the camera parameters are not provided in the COCO dataset, we choose a general focal length (\ie 1400) and assume the pitch angle of the camera is zero. We can see that our approach not only obtains accurate 3D pose for each person in the image but also estimates their absolute depth correctly in the 3D space. In particular, the model is robust to pose and background variations. For instance, in the baseball example, we get correct depth estimate for the person in sitting posture. Fig. \ref{fig:experiment_quality}(b) and (c) show some typical failure cases. In Fig. \ref{fig:experiment_quality}(b), the man in the red circle is occluded and truncated, while in Fig. \ref{fig:experiment_quality}(c), the little girl is much shorter than the people in the training dataset. In the future, to address the second issue, we will study the possibility of adding another parallel branch in the 2D network to estimate a correction factor to refine the estimated depth of each person.

\section{Conclusion}
\label{sec:conclusion}
In this work, we present a systematic study of the generalization problem of 3D pose estimation. We hope the study can inspire future works to consider the generalization aspect when designing and evaluating new models. Then we present \emph{VirtualPose}, an approach for absolute 3D human pose estimation, which does not require paired images and 3D poses for training. In particular, part of the network is trained on abundant 2D datasets such as COCO and the rest are trained on synthetic datasets. The decoupling strategy helps the approach avoid over-fitting to small training data. As a result, it can be flexibly adapted to a new environment with minimal human effort. Our method achieves significantly better results than the existing methods on the benchmark datasets especially when the training and test datasets have different distributions.

\subsection{Future Work}
First, in the current implementation, AGR only consists of 2D joint heatmaps and person-level bounding boxes which are powerful to recover the absolute positions of the root joints. But we can actually explore finer-grained AGRs such as segmentation maps and  occlusion relationship maps which can benefit more to relative pose estimation. Second, we only used the 3D poses from the Panoptic and MuCo-3DHP dataset for generating AGR training data. But we can actually generate more poses by manipulating joint angles which may further improve the results.

\paragraph{\bf Acknowledgement.}  This work was supported in part by MOST-2018AAA0102004 and NSFC-62061136001.

\clearpage
%
%
\bibliographystyle{splncs04}
\bibliography{egbib}

\end{document}